\documentclass[letterpaper, 10 pt, conference]{ieeeconf}
\IEEEoverridecommandlockouts                              

\usepackage{cite}
\usepackage{multirow}
\usepackage{algorithm}  	
\usepackage{algorithmic}
\usepackage{graphicx}
\usepackage{times}
\usepackage{amsmath}
\usepackage{amssymb}
\usepackage{subfigure}
\usepackage{graphics} 
\usepackage[utf8]{inputenc}
\usepackage{boldline}
\usepackage{multicol}
\usepackage[pdfstartview=FitH,bookmarksnumbered=true,colorlinks,bookmarksopen=true]{hyperref}
\newcommand{\etal}{\textit{et al}.~}

\title{\LARGE \bf
Vision-based Teleoperation of Shadow Dexterous
Hand \\using End-to-End Deep Neural Network
}

\author{Shuang Li$^{1\dagger}$, Xiaojian Ma$^{2\dagger}$, Hongzhuo Liang$^{1}$, Michael G\"{o}rner$^{1}$, Philipp Ruppel$^{1}$,\\
 Bin Fang$^{2*}$, Fuchun Sun$^{2}$, Jianwei Zhang$^{1}$
\thanks{$\dagger$These two authors contributed equally. This work was done when Shuang Li was visiting Tsinghua University.
}
\thanks{$^{1}$TAMS (Technical Aspects of Multimodal Systems), Department of Informatics, Universit\"{a}t Hamburg}
\thanks{$^{2}$Tsinghua National Laboratory for Information Science and Technology (TNList), State Key Lab on Intelligent Technology and Systems, Department of Computer Science and Technology, Tsinghua University}
\thanks{*Corresponding author to provide e-mail: fangbin@mail.tsinghua.edu.cn}%
}
\begin{document}

\maketitle
\thispagestyle{empty}
\pagestyle{empty}

\begin{abstract}
In this paper, we present TeachNet, a novel neural network architecture for intuitive and markerless vision-based teleoperation of dexterous robotic hands.
Robot joint angles are directly generated from depth images of the human hand that produce visually similar robot hand poses in an end-to-end fashion. The special structure of TeachNet, combined with a consistency loss function, handles the differences in appearance and anatomy between human and robotic hands. A synchronized human-robot training set is generated from an existing dataset of labeled depth images of the human hand and simulated depth images of a robotic hand. The final training set includes 400K pairwise depth images and joint angles of a Shadow C6 robotic hand. 
The network evaluation results verify the superiority of TeachNet, especially regarding the high-precision condition. 
Imitation experiments and grasp tasks teleoperated by novice users demonstrate that TeachNet is more reliable and faster than the state-of-the-art vision-based teleoperation method.
\end{abstract}

\section{Introduction}
Robotic dexterous hands provide a promising base for supplanting human hands in the execution of tedious and dangerous tasks. When autonomous manipulation of dexterous hands handles complex perception
, teleoperation is superior to intelligent programming when it comes to taking fast decisions and dealing with corner cases.

Unlike contacting or wearable device-based teleoperation, markerless vision-based teleoperation \cite{tele2007advances} offers the advantages of showing natural human-limb motions and of being less invasive. Analytical vision-based teleoperation falls into two categories: model- and appearance-based approaches. Model-based approaches \cite{markerless2gripper, hand-arm} provide continuous solutions but are computationally costly and typically depend on the availability of a multicamera system \cite{model-based-vision-teleop}. Conversely, appearance-based approaches \cite{phdthesis, gesture_teleop} recognize a discrete number of hand poses that correspond typically to the method’s training set without high computational cost and hardware complexity. Recently, an increasing number of researchers have been focusing on the data-driven vision-based teleoperation methods which get the 3D hand pose or recognize the class of hand gestures using first the deep convolutional neural network (CNN) then mapping the locations or the corresponding poses to the robot. However, all these solutions not only strongly depend on the accuracy of the hand pose estimation or the classification but also suffer the time cost of post-processing.

We instead seek to take a noisy depth image of the human hand as input and produce joint angles of the robot hand as output by training a deep CNN. The end-to-end vision-based teleoperation can be a natural and an intuitive way to manipulate the remote robot and is user-friendly to the novice
teleoperators. Therefore, it is essential to design an efficient network which could learn the corresponding robot pose feature in human pose space. Since the end-to-end method depends on massive human-robot teleoperation pairings, we aim to explore an efficient method which collects synchronized hand data both for the robot and the human. 
\begin{figure}[t]
	\includegraphics[width=0.35\textheight]{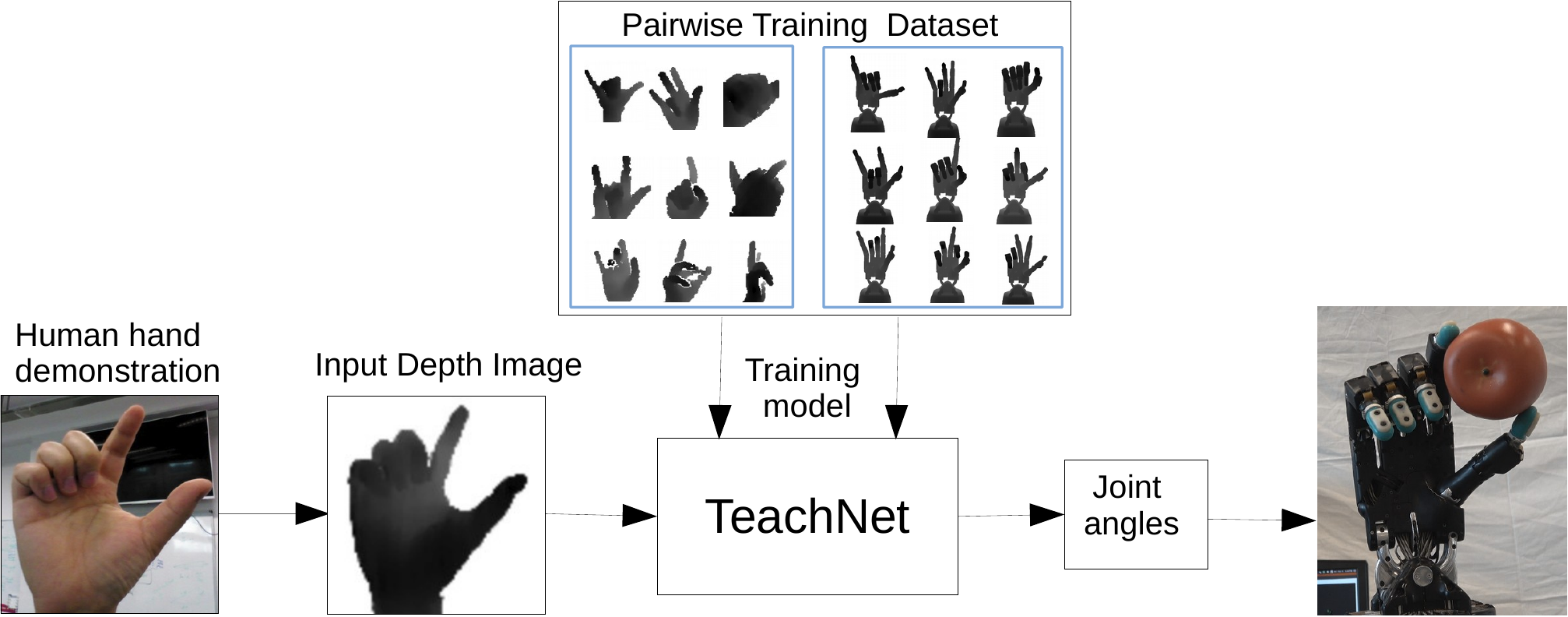}	
	\caption{Our vision-based teleoperation architecture. 
		(Center) TeachNet is trained offline to predict robot joint angles from depth images of a human hand using our 400k pairwise human-robot hand dataset.
		(Left)Depth images of the operator's hand are captured by a depth camera then feed to TeachNet.
		(Right) The joint angles produced by TeachNet are executed on the robot to imitate the operator's hand pose.
		}
	\label{intro}
	\vskip -0.15in
\end{figure}

In this paper, we present a novel scheme for teleoperating the Shadow dexterous hand based on a single depth image (see Fig. \ref{intro}).
Our primary contributions are: 1) We propose an end-to-end teacher-student network(TeachNet), which learns the kinematic mappings between the robot and the human hand. 2) We build a pairwise human-robot hand dataset that includes pairs of depth images in the same gesture, as well as corresponding joint angles of the robot hand. 3) We design an optimized mapping method that matches the Cartesian position and the link direction of shadow hand from the human hand pose and properly takes into account possible self-collisions.

During the network evaluation, TeachNet achieves higher accuracy and lower error compared to other end-to-end baselines.
As we illustrated in our robotic experiments, our method allows the Shadow robot to imitate human gestures and to finish the grasp tasks significantly faster than state-of-the-art data-driven vision-based teleoperation.

\section{Related Work}
\noindent
\textbf{Markerless Vision-Based Teleoperation.}
Human teleoperation of robots has usually been implemented through contacting devices such as tracking sensors \cite{SCHUNKS5FHteleop}, gloves instrumented with angle sensors \cite{fang2015robotic, fang20183d}, inertial sensors \cite{miller2004motion} and joysticks \cite{cho2010teleoperation}.
Stanton \etal \cite{humanoid2012pairs} suggest an end-to-end teleoperation on a 23 degree of freedom (DOF) robot by training a feed-forward neural network for each DOF of the robot to learn the mapping between sensor data from the motion capture suit and the angular position of the robot actuator to which each neural network is allocated. 
However, wearable devices are customized for a certain size range of the human hand or the human body, and contacting methods may hinder natural human-limb motion.
Compared to these methods, markerless vision-based teleoperation is less invasive and performs natural and comfortable gestures.

Visual model-based methods, such as \cite{markerless2gripper, hand-arm}, compute continuous 3D positions and orientations of thumb and index finger from segmented images based on a camera system and control a parallel jaw gripper mounted on a six-axis robot arm. 
Romero \cite{phdthesis} classifies human grasps into grasp classes and approaches based on human hand images then maps them to a discrete set of corresponding robot grasp classes following the external observation paradigm.

Compared to analytical methods, data-driven techniques place more weight on object representation and perceptual processing, e.g., feature extraction, object recognition or classification and pose estimation. Michel \etal \cite{markerless_humanpose} provide a teleoperate method for a NAO humanoid robot that tracks human body motion from markerless visual observations then calculates the inverse kinematics process. But this method does not consider the physical constraints and joint limits of the robots, so it easily generates the poses that the robot cannot reach.
Nevertheless, these methods strongly depend on the accuracy of the hand pose estimation or the classification and lose much time for post-processing.
In this work, we aim to design an end-to-end vision-based CNN which generates continuous robot poses and provides the fast and intuitive experience of teleoperation.

\noindent
\textbf{Depth-Based 3D Hand Pose Estimation.}
3D hand pose estimation typically is one of the essential research fields in vision-based teleoperation. Although the field of 3D hand pose estimation has advanced rapidly, isolated 3D hand pose estimation only achieves low mean errors (10 mm) in the view point range of [70, 120] degrees \cite{depth_survey}. According to the representation of the output pose, the 3D hand pose estimation methods consist of detection- and regression-based methods. Detection-based methods \cite{v2v} give the probability density map for each joint, while regression-based methods \cite{deepprior++, ren} directly map the depth image to the joint locations or the joint angles of a hand model. 
Regardless of whom the output joint pose belongs to, the regression-based network is similar to our end-to-end network.

\begin{figure}[ht]
	\centering
	\includegraphics[width=0.35\textheight]{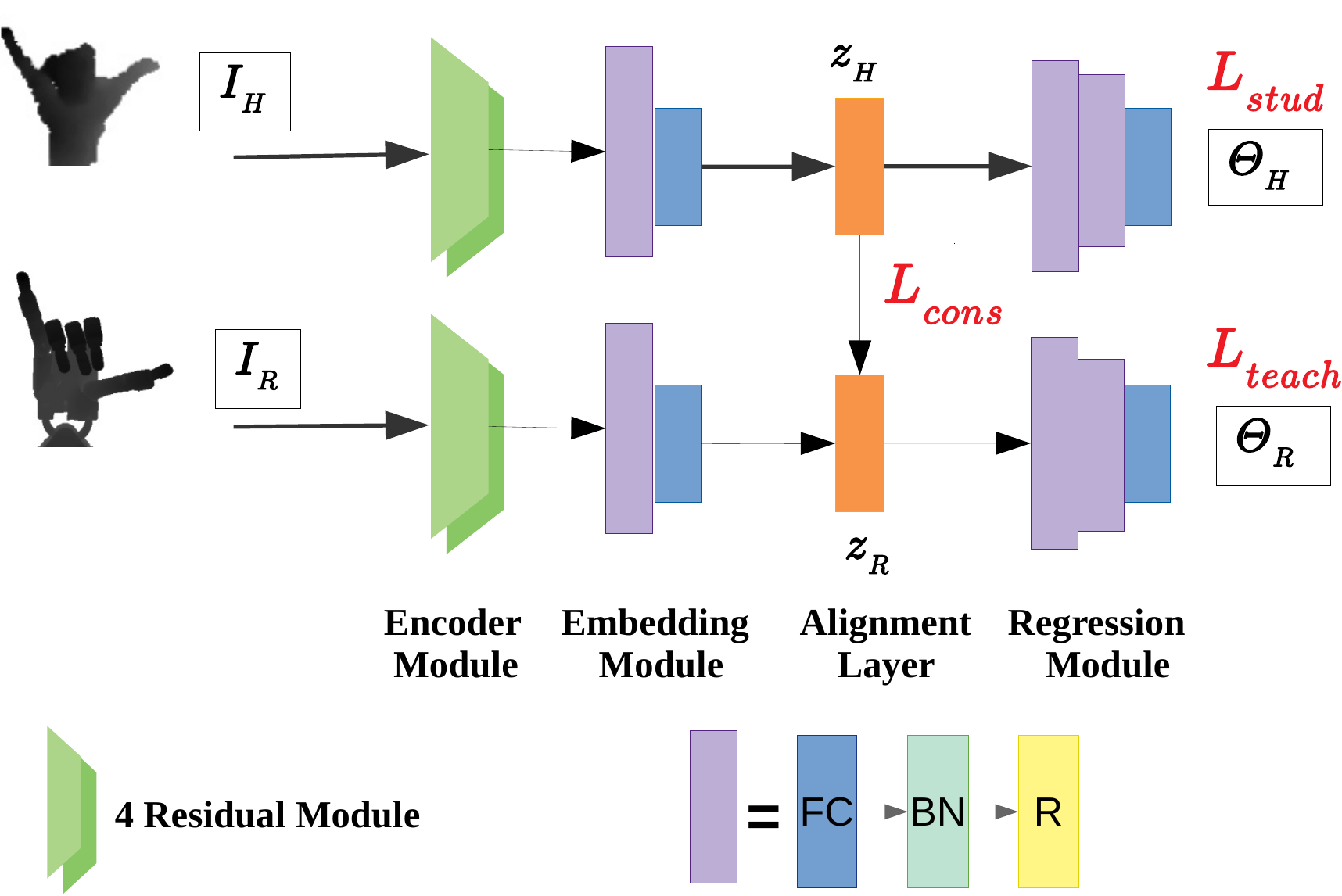}	
	\caption{TeachNet Architecture. Top: human branch, Bottom: robot branch. The input depth images $I_H$ and $I_R$ are fed to the corresponding branch that predicts the robot joint angels $\Theta_H$, $\Theta_R$. The residual module is a convolutional neural network with a similar architecture as ResNet~\cite{resnet}. FC denotes a fully-connected layer, BN denotes a batch normalization layer,
    R denotes a Rectified Linear Unit.
	}
	\label{teachnet}
	\vskip -0.15in
\end{figure}

\noindent
\textbf{Master-Slave Pairing in Teleoperation.}
To learn the pose feature of the robot from the images of the human hand, we have to consider how to get a vast number of the human-robot pairings.
Prior work in \cite{humanoid2012pairs, gussian_pairs, tcn} acquired the master-slave pairings by demanding a human operator to imitate the robot motion synchronously. The pairing data is costly to collect like this and typically comes with noisy correspondences.
Also, there is no longer an exact correspondence between the human and the robot because physiological differences make the imitation non-trivial and subjective to the imitator. 
In fact, the robot state is more accessible and is relatively stable concerning the human hand, and there are many existing human hand datasets. 
Since training on real images may require significant data collection time, an alternative approach is to learn on simulated images and to adapt the representation to real data \cite{openai}.
We propose a novel criterion of generating human-robot pairing from these results by using an existing dataset of labeled human hand depth images, manipulating the robot and recording corresponding joint angles and images in simulation, and performing extensive evaluations on a physical robot.

\noindent
\textbf{Teleoperation Mapping Methods.}
Conventional teleoperation mapping methods are divided into three main categories: joint mapping which is useful for power grasps \cite{joint_teleop}, fingertip mapping which is suitable for precision grasps \cite{optimizedfingertip}, and pose mapping which interprets the function of the human grasp rather than replicating hand position \cite{meeker2018intuitive}.
However, in most cases considering only one type of mapping method is not enough \cite{hybrid_teleop}. For example, fingertip mapping neglects the position and orientation of the phalanges and does not consider the special mechanical difference between the slave and the master.

\section{Teacher-Student Network}
Solving joint regression problems directly from human images is quite challenging because the robot hand and the human hand occupy two different domains. Specifically, imagine that we have image $I_R$ of a robotic hand and image $I_H$ of a human hand, while the robotic hand in the image acts exactly the same as the human hand. The problem of mapping the human hand image to the corresponding robotic joint could be formulated as below:
\begin{align}\begin{split}\label{human_regress}
    & f_{feat}: I_H \in \mathbb{R}^2 \rightarrow z_{pose} \\
    & f_{regress}: z_{pose} \rightarrow \Theta\text{.}   
\end{split}\end{align}

To better process the geometric information in the input depth image and the complex constraints on joint regression, we adopt an encode-decode style deep neural network. The upper branch in Fig. \ref{teachnet} illustrates the network architecture we used. However, 
the human hand and shadow hand basically come from different domains, thus it could be difficult for $f_{feat}$ to learn an appropriate latent feature $z_{pose}$ in pose space. In contrast, the mapping from $I_R$ to joint target $\Theta$ will be more natural as it is exactly a well-defined hand pose estimation problem. Intuitively, we believe that for a paired human and robotic image, their latent pose features $z_{pose}$ should be encouraged to be consistent as they represent the same hand pose and will be finally mapped to the same joint target. Also, based on the observation that the mapping from $I_R$ to $\Theta$ performs better than $I_H $ (these preliminary results can be found in Fig. \ref{angle_eval}), the encoder $f_{feat}$ of $I_R$ could extract better pose features, which could significantly improve the regression results of the decoder.

With these considerations above, we propose a novel teacher-student network (TeachNet) to tackle the vision-based teleoperation problem~\eqref{human_regress} in an end-to-end fashion. 
TeachNet consists of two branches, the robot branch which plays the role of a teacher and the human branch as the student.

\noindent
\textbf{Joint angle loss.} Each branch is supervised with a mean squared error (MSE) loss $\mathcal{L}_{ang}$ :
	\begin{equation}
	\label{angloss}
	\mathcal{L}_{ang} = \|\Theta - J \|^2\text{,}
	\end{equation}
where 
$J$ is the groundtruth joint angles.

Besides the encoder-decoder structure that maps the input depth image to joint prediction, we define a consistency loss $\mathcal{L}_{cons}$ between two latent features $z_{H}$ and $z_{R}$ to exploit the geometrical resemblance between human hands and the robotic hand. Therefore, $\mathcal{L}_{cons}$ forces the human branch to be supervised by a pose space shared with the robot branch. To explore the most effective aligning mechanism, we design two kinds of consistency losses and two different aligning positions:

\noindent
\textbf{Hard consistency loss.}
The most intuitive mechanism for feature alignment would be providing an extra mean squared error (MSE) loss over the latent features of these two branches:
	\begin{align}\label{hard_loss}
	 \mathcal{L}_{cons\_h} = \|z_{H} - z_{R} \|_2\text{.}
	\end{align}

\noindent
\textbf{Soft consistency loss.} 
Sometimes, \eqref{hard_loss} could distract the network from learning hand pose representations especially in the early training stage. Inspired by~\cite{villegas2018neural}, we feed $z_{H}$ and $z_{R}$ into a discriminator network $D$~\cite{gan} to compute a \textit{realism score} for \textit{real} and \textit{fake} pose features. The soft consistency loss is basically the negative of this score:
\begin{align}\label{soft_loss}
\mathcal{L}_{cons\_s} = \log\left(1-D(z_{H})\right)\text{.}
\end{align}

As for the aligning position, we propose \textit{early teaching} and \textit{late teaching} respectively. For the former, we put the alignment layer after the encoder and embedding module, while in the latter the alignment layer is positioned on the last but one layer of the whole model (which means that the regression module will only contain one layer).

In the following, we will refer to early teaching by $\mathcal{L}_{cons\_s}$  as Teach Soft-Early, late teaching by $\mathcal{L}_{cons\_s}$ as Teach Soft-Late, early teaching by $\mathcal{L}_{cons\_h}$ as Teach Hard-Early, and late teaching by $\mathcal{L}_{cons\_h}$ as Teach Hard-Late. 

We also introduce an auxiliary loss to further improve our teleoperation model:

\noindent
\textbf{Physical loss.} The physical loss $\mathcal{L}_{phy}$ which enforces the physical constraints and joint limits is defined by:
\begin{equation}
\label{phyloss}
\mathcal{L}_{phy}(\Theta)=\sum\limits_i[\max(0, (\theta_{max} - \Theta_i)) + \max(0, (\Theta_i - \Theta_{min}))]\text{.}
\end{equation}
Overall, the complete training objective for each branch is:
\begin{equation}
\label{teachloss}
\mathcal{L}_{teach}(\Theta)=\mathcal{L}_{ang} + \mathcal{L}_{phy}
\end{equation}
\begin{equation}
\label{studloss}
\mathcal{L}_{stud}(\Theta)=\mathcal{L}_{ang} + \alpha * \mathcal{L}_{cons} + \mathcal{L}_{phy}\text{,}
\end{equation}
where $\alpha=1$ for hard consistency loss and $\alpha=0.1$ for soft consistency loss.

\section{Dataset Generation}
\label{dataset}
\begin{figure}[!t]
	\centering
	\includegraphics[width=0.35\textheight]{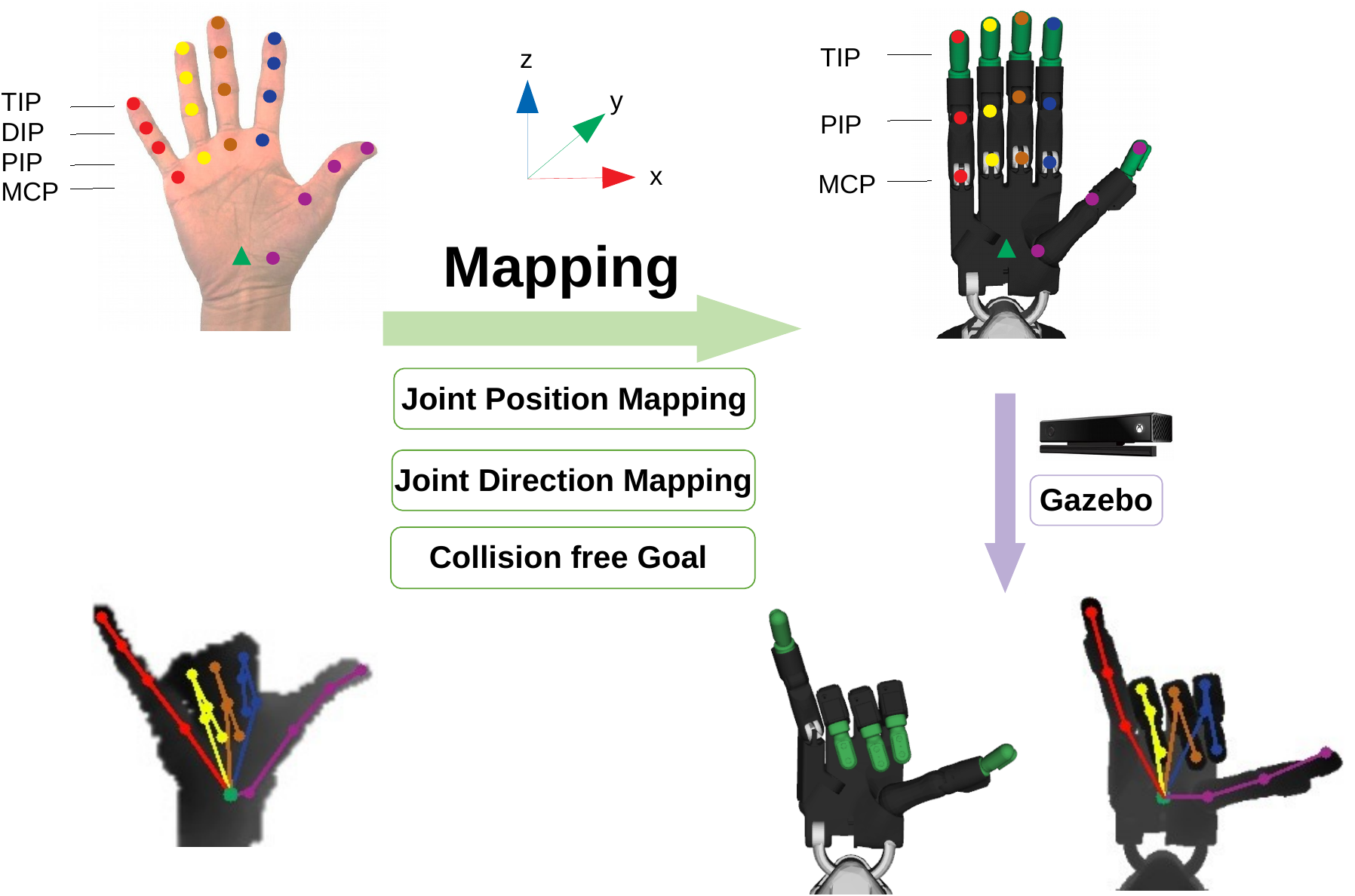}
	\caption{Pipeline for dataset generation. (Top left) The human hand model has 21 joints and moves with 31 degrees of freedom in the BigHand2.2M dataset. (Bottom left) A depth image example from the BigHand2.2M dataset. (Middle) Optimized mapping method from the human hand to the Shadow hand. (Top right) The Shadow hand with BioTac sensors has 24 joints and moves with 19 degrees of freedom. (Bottom right) The corresponding RGB and depth images of Shadow gestures obtained from Gazebo. The colored circles denote the joint keypoint positions on the hand, and the green triangles denote the common reference frame $F$.}
	\label{map}
	\vskip -0.15in
\end{figure}

Training the TeachNet which learns the kinematic mapping between the human hand and the robot hand strongly relies on a massive dataset with human-robot pairings.
We achieve this by using the off-the-shelf human hand dataset BigHand2.2M Dataset \cite{bighand2} and an optimized mapping method using the pipeline of Fig. \ref{map}.
With this pipeline, we collect a training dataset that contains 400K pairs of simulated robot depth images and human hand depth images, with corresponding robot joint angles and poses.

\subsection{Human-Robot Hand Kinematic Configuration}
The Shadow Dexterous Hand \cite{hand2013e1} used in this work is motor-controlled and equipped with 5 fingers, and its kinematic chain is shown in the right side of Fig. \ref{error}. Each of these fingers has a BioTac tactile sensor attached which replaces the last phalanx and the controllability of the last joint. 
Each finger has four joints, the distal, middle, proximal, and the metacarpal joint, but the first joint of each finger is stiff. 
The little finger and the thumb are provided with an extra joint for holding the objects.
Summed up, this makes 17 DOF plus two in the wrist makes 19.

In contrast to the robot hand, the human hand model from BigHand2.2M dataset has 21 joints and can move with 31 DOF, as shown in Fig. \ref{map}. 
The main kinematic differences between the robot hand and the human hand are the limited angle ranges of the robot joints and the structure of the wrist joints. 
Simplifying the dissimilarity between the Shadow hand and the human hand, two wrist joints of the Shadow at $0$ rad are fixed and only 15 joint keypoints which are TIP, PIP, MCP in each finger of the Shadow are considered.

\subsection{Optimized Mapping Method}
Effectively mapping the robot pose from the human hand pose plays a significant role in our training dataset.
In order to imitate the human hand pose, we propose an optimized mapping method integrating position mapping,  orientation mapping and properly taking into account possible self-collisions.

Firstly, we use the common reference frame $F$ located at the human wrist joint and 34mm above the z-axis of the robot wrist joint. Note that 34mm is the height from the wrist joint to the base joint of the thumb. 
These locations are chosen because they lie at locations of high kinematic similarity.
Secondly, we enforce position mapping to the fingertips with a strong weight $\omega_{pf}$ and to PIP joints with minor weight $\omega_{pp}$. 
Thirdly, direction mapping with weight $\omega_{d}$ is applied to five proximal phalanges and distal phalanges of thumb. 
In our dataset, we set $\{\omega_{pf}, \omega_{pp}, \omega_{d}\} = \{1, 0.2 ,0.2\}$.

Taking advantage of BioIK solver \cite{bioik} to determine the robot joint angles $\Theta \in R^{17}$, the robot execute movements in Gazebo and check self-collision by MoveIt. In case BioIK gives a self-collision output, we define a cost function $F_{cost}$ which measures the distance between two links

\begin{figure}[htb]
	\centering
	\includegraphics[width=0.16\textheight]{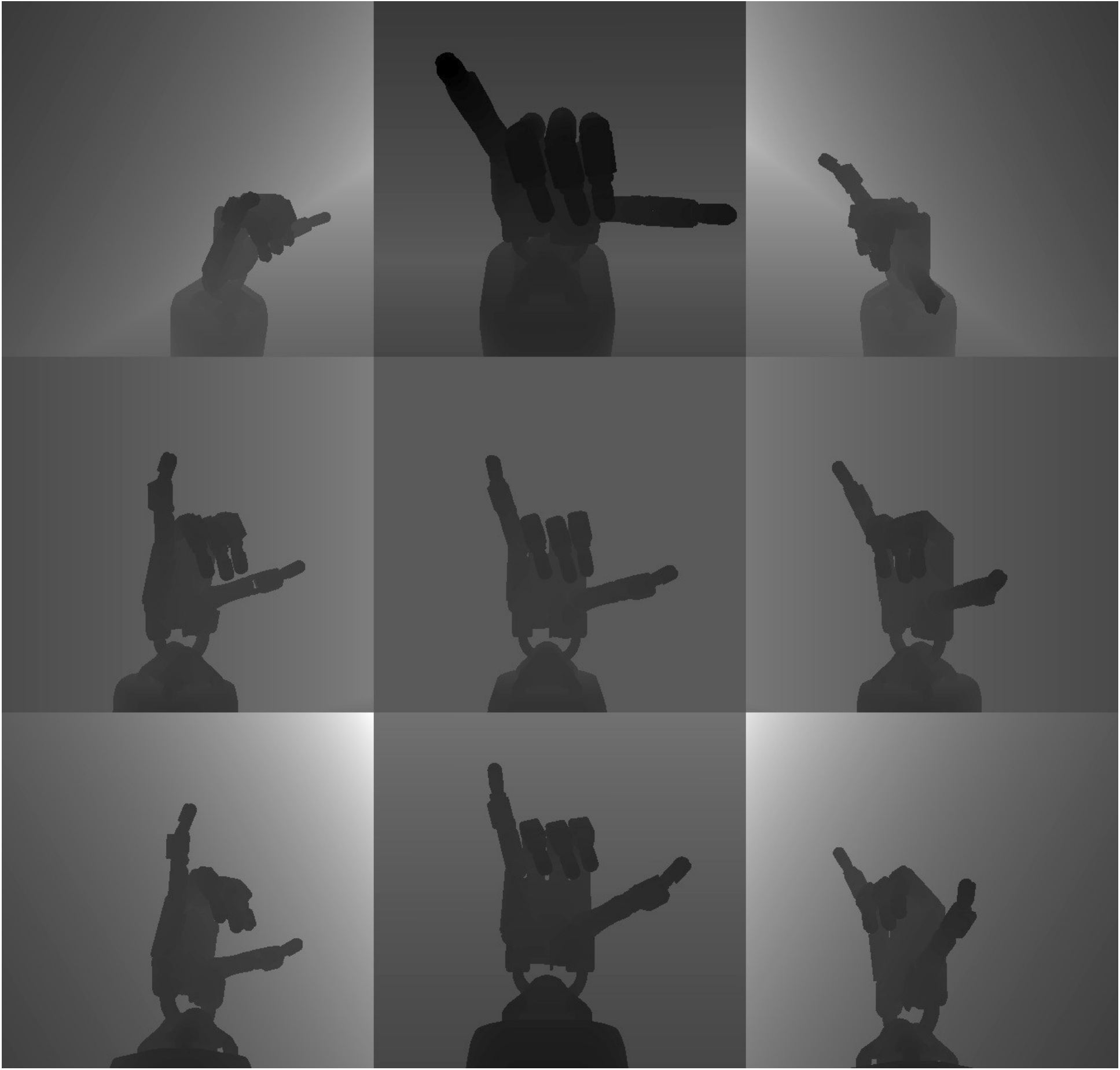}
	\caption{The Shadow depth images from nine viewpoints corresponding to one human gesture in our dataset.}
	\label{dataset_result}
	\vskip -0.15in
\end{figure}

\begin{equation}
\label{collision}
F_{cost}=\max(0, R\_col - \|P_i - P_j \|^2)\text{,}
\end{equation}
where $P_i$, $P_j$ respectively denote the position of link i, link j, $R\_col$ is the minimum collision free radius between two links.

Considering the BigHand2.2M dataset spans a wide range of observing viewpoints to the human hand, it is indispensable to increase the diversity of viewpoints of the robot data. Thus we collect visual samples of the robot through nine simulated depth cameras with different observing positions in Gazebo and record nine depth image for each pose simultaneously. As an example, in Fig.~\ref{dataset_result} we present the depth images of the robot from nine viewpoints corresponding to the human hand pose at the bottom left in Fig.~\ref{map}.


\section{Experiment} 
\subsection{TeachNet Evaluation}
\label{network baseline}
We examined whether the TeachNet could learn more indicative visual representations that were the kinematic structure of the human hand. The proposed TeachNet was evaluated on our paired dataset with the following experiments: 1) To explore the appropriate position of the alignment layer and the proper align method, we compared the proposed four network structures: Teach Soft-Early, Teach Soft-Late, Teach Hard-Early, and Teach Hard-Late. 2) To validate the significance of the alignment layer, we designed an ablation analysis by removing consistency loss $\mathcal{L}_{cons}$ and separately training the single human branch and the single robot branch. We respectively refer to these two baselines as Single Human and Single Robot. 3) We compared our end-to-end method with the data-driven vision-based teleoperation method which mapped the position of the robot from the human joint locations based on the 3D hand estimation. We refer to this baseline as HandIK solution. There were three evaluation metrics used in this work: 1) the fraction of frames whose maximum/average joint angle errors are below a threshold; 2) the fraction of frames whose maximum/average joint distance errors are below a threshold; 3) the average angle error over all angles in $\Theta$.

The input depth images of all network evaluations were extracted from the raw depth image as a fixed-size cube around the hand and resized to $100 \times 100$. 
Note that although we have nine views of Shadow images which correspond to one human pose, during the training process of the TeachNet we randomly chose one view of Shadow images to feed into the robot branch. 
For the HandIK solution, we trained the DeepPrior++ network on our dataset, and the reason we chose DeepPrior++ was that its architecture was similar to the single branch of TeachNet. 
We obtained the $21 \times 3$ human joint locations from DeepPrior++ then used the same mapping method in section \ref{dataset} to acquire the joint angles of the Shadow hand.

\begin{figure*}[htbp]
	\centering
	\includegraphics[width=1.0\textwidth]{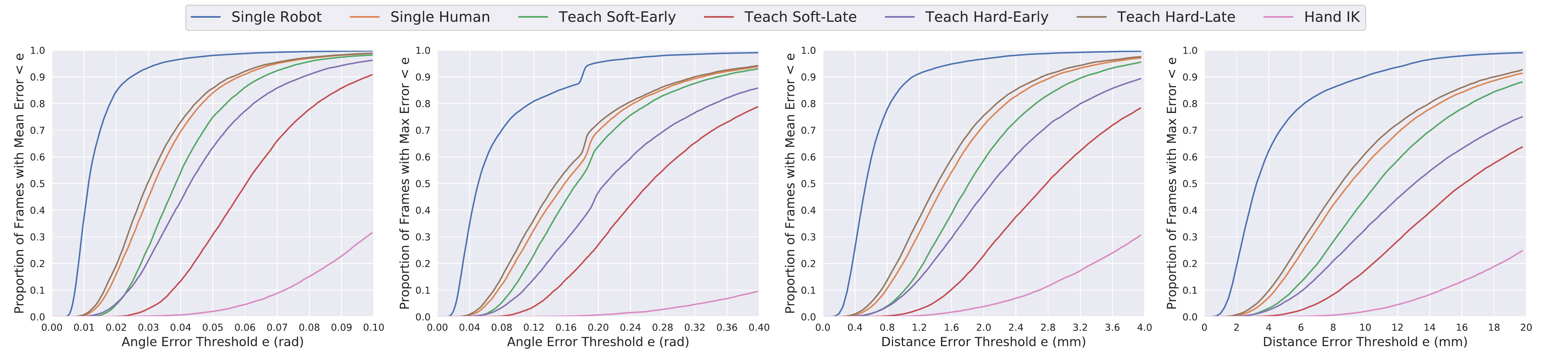}
	\caption{The fraction of frames whose maximum/average joint angle/distance error are below a threshold between Teach Hard-Late approach and different baselines on our test dataset.
	These show that Teach Hard-Late approach has the best accuracy over all evaluation metrics.
	}
	\vskip -0.15in
	\label{angle_eval}
\end{figure*}

\begin{figure*}[htbp]
	\centering
	\includegraphics[width=\textwidth]{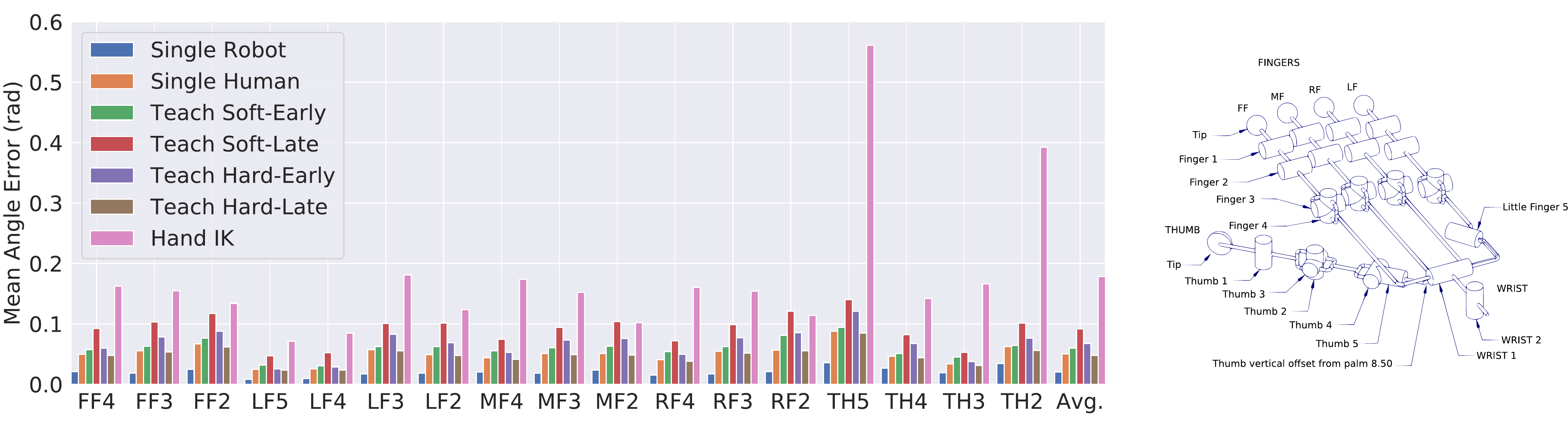}
	\caption{(Left) Comparison of average angle error on the individual joint between the Teach Hard-Late approach and different baselines on our test dataset. FF means the first finger, LF means the little finger, MF means the middle finger, RF means the ring finger, TH means the thumb. (Right) The kinematic chain of the Shadow hand. In this work, joint 1 of each finger is stiff.}
	\label{error}
	\vskip -0.15in
\end{figure*}


\begin{table}[ht]
	\centering
	\caption{Accuracy under High-Precision Conditions}
	\vskip -0.15in
	\begin{tabular}{cccc}
		\multicolumn{3}{c}{}\\
		\hlineB{2}
		Max Err.          & Single Human & Teach Soft-Early & Teach Soft-Late \\ \hline
		0.1 rad  & 21.24\%        & 12.31\%            & 12.77\%           \\
		0.15 rad & 45.57\%        & 38.06\%            & 10.37\%           \\
		0.2 rad  & 69.08\%        & 63.18\%            & 26.16\%           \\ \hlineB{2}
		
		\multicolumn{3}{c}{}\\
		\hlineB{2}
		Max Err.          & Teach Hard-Early & Teach Hard-Late & Hand IK \\ \hline
		0.1 rad  & 7.40\%        & \textbf{24.63\%}            & 0.00\%           \\
		0.15 rad & 24.67\%       & \textbf{50.11\%}            & 0.14\%           \\
		0.2 rad  & 45.63\%       & \textbf{72.04\%}            & 0.62\%           \\ \hlineB{2}
	\end{tabular}
	\label{tab:high_precision_acc}
	\vskip -0.2in
\end{table}

The comparative results, shown in Fig. \ref{angle_eval} and Fig. \ref{error}, indicate that the Single Robot method is the best concerning all evaluation metrics and has the capability of the training "supervisor". 
Meanwhile, the Teach Hard-Late method outperforms the other baselines, which verifies that the single human branch is enhanced through an additional consistency loss. 
Especially regarding the high-precision condition, only the Teach Hard-Late approach shows an average $3.63\%$ improvement of the accuracy below a maximum joint angle which is higher than that of the Single Human method (Table \ref{tab:high_precision_acc}). We refer that the later feature space $f_{feat}$ of the depth images contains more useful information and the MSE method displays the stronger supervision in our case.
And the regression-based HandIK method shows the worst performance among our three metrics. The unsatisfying outcome of the HandIK solution is not only down to our network giving a better representation of the hand feature but also due to the fact that this method does not consider the kinematic structure and the special limitation of the robot. Furthermore, direct joint angle regression should have decent accuracy on angles since that is the learning objective. The missing $L_{phy}$ also gives rise to poor accuracy.

Moreover, Fig. \ref{error} demonstrates that the second joint, the third joint and the base joint of the thumb are harder to be learned.
These results are mainly because that 1) The fixed distal joints of the robot in our work affect the accuracy of its second joint and third joint. 2) these types of joints have a bigger joint range than other joints, especially the base joint of the thumb. 3) there is a big discrepancy between the human thumb and the Shadow thumb.

\subsection{Robotic Experiments}
To verify the reliability and intuitiveness of our method, real-world experiments were performed with five grown-up subjects. The slave hand of our teleoperation system is the Shadow dexterous hand where the first joint of each finger is fixed. The depth sensor is the Intel RealSense F200 depth sensor which is suitable for close-range tracking. The poses of the teleoperators' right hand are limited to the viewpoint range of [70$^{\circ}$, 120$^{\circ}$] and the distance range of [15mm, 40mm] from the camera. Since the vision-based teleoperation is susceptible to the light situation, all the experiments were carried out under a uniform and bright light source as much as possible. The average computation time of the Teach Hard-Late method is 0.1051s(Alienware15 with Intel Core i7-4720HQ CPU). Code and video are available at \href{https://github.com/TAMS-Group/TeachNet_Teleoperation}{https://github.com/TAMS-Group/TeachNet\_Teleoperation}.

\subsubsection{Simulation Experiments}
The five novice teleoperators stood in front of the depth sensor and performed 0-9 in American sign language and random common gestures in a disordered way, then teleoperated the simulated Shadow robot.
The operators did not need to know the control mechanism of the robot and naturally implemented the experiment.

\begin{figure}[ht]
	\centering
	\subfigure[Successful teleoperation results]{
		\begin{minipage}{0.45\textwidth}
			\centering\label{good}
			\includegraphics[width=1\textwidth]{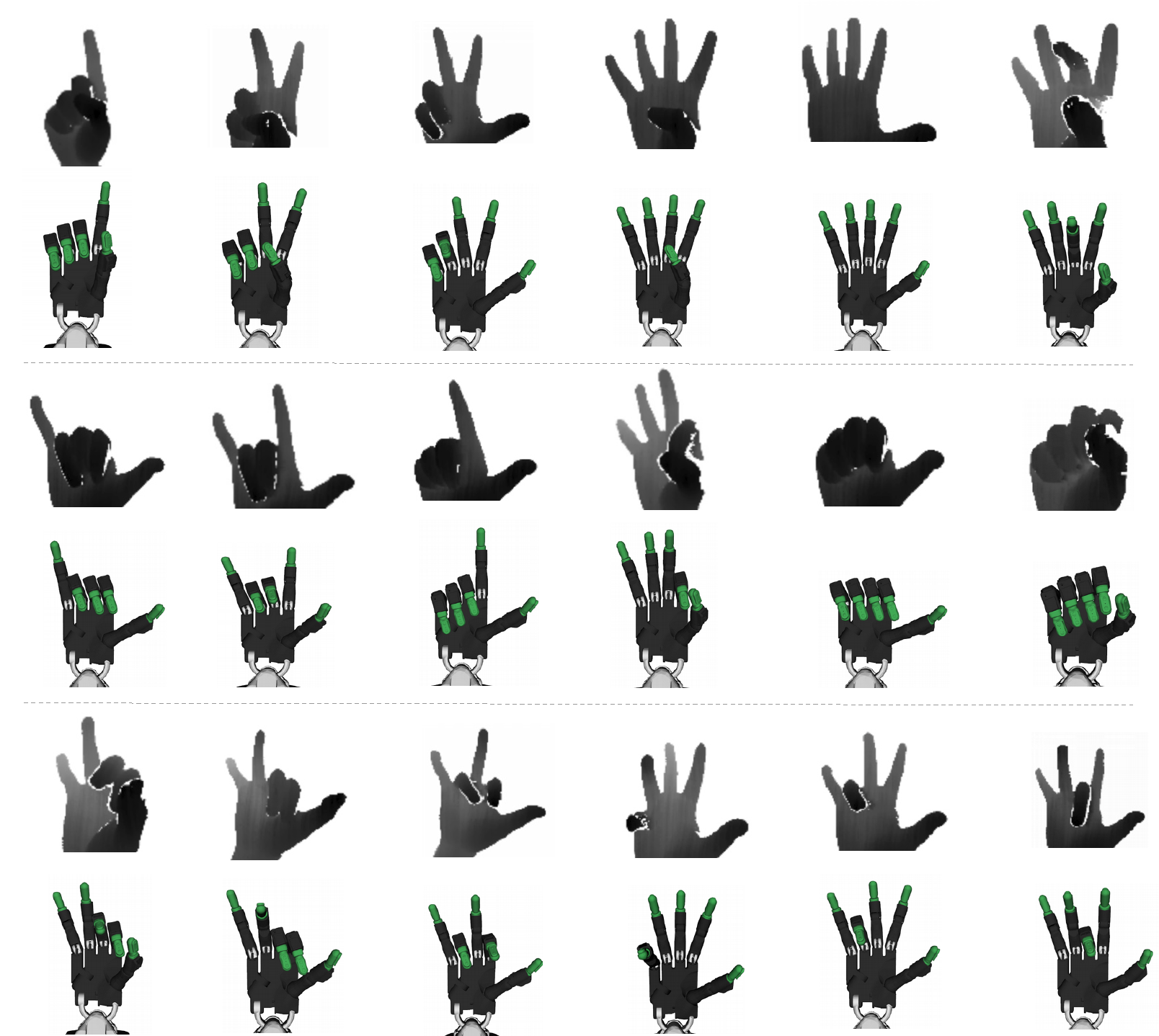}
		\end{minipage}}
		\subfigure[Failed teleoperation results]{
			\begin{minipage}{0.36\textwidth}
				\centering\label{bad}
				\includegraphics[width=1\textwidth]{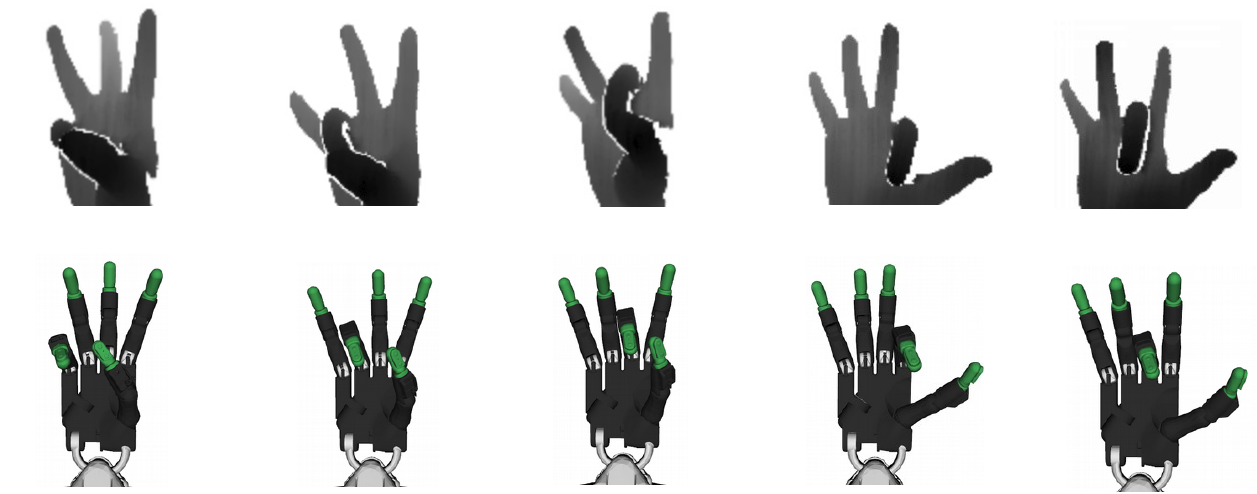}
			\end{minipage}}
		\caption{Teleoperation results using the Shadow hand on  real-world data.}
		\label{simulation_result}
\end{figure}

Qualitative results of teleoperation by the Teach Hard-Late method are illustrated in Fig. \ref{simulation_result}.
We can see that the Shadow hand vividly imitates human gestures of different size of human hands. These experiments demonstrate that the TeachNet enables a robot hand to perform continuous, online imitation of human hand without explicitly specifying any joint-level correspondences. Owing to the fact that we fixed two wrist joints of the Shadow hand, we did not care if the depth sensor captures the teleoperator's wrist.

However, visible errors occurred mainly with the second joint, the third joint of the fingers, and the base joint of the thumb, probably caused by the special kinematic structure of the slave, occlusions and the uncertain lighting conditions.

\subsubsection{Manipulation Experiments}
We compared the Teach Hard-Late method with the deep prior++ HandIK method on a slave robot. To simplify our experiments, we set the control mode of the robot to be the trajectory control within a proper maximum force for each joint.
We used time to complete an in-hand grasp and release task as a metric for usability.

We placed a series of objects in the slave hand which was in the \textit{open} pose one at a time to facilitate easier grasping with the robotic fingers and asked subjects to grasp them up then release them. The objects 
used for the grasp and release tasks were: a water bottle, a small mug, a plastic banana, a cylinder can, and a plastic apple. 
We required the operators to use power grasp for the water bottle and the mug, and to use precision grasp for other objects. If the user did not complete the task in four minutes, they were considered to be unable to grasp the object. 

Table \ref{grasp} numerically shows the average time a novice took to grasp an object using each of the control methods. We find that the low accuracy, especially for the thumb, and the post-processing of the HandIK solution results in a longer time to finish the task. The users needed to open the thumb first then perform proper grasp action, so HandIK solution shows worse performance for the objects with a big diameter.
Besides that, grasping the banana took the longest time on our method because the long and narrowly shaped object needed more precious fingertip position.

\begin{table}[ht]
    \centering
    \caption{Average Time a Novice
Took to Grasp and Release an Object}
    \begin{tabular}{ccccccc}\hlineB{2}
    Methods & Bottle & Mug & Banana & Can & Apple & Average\\ \hline
    Hand IK & 44.15& 46.32& 35.78& 25.50& 30.22& 36.394\\
    Ours & 23.67& 18.82& 25.80& 19.75& 15.60&  20.728\\
      \hlineB{2}
    \end{tabular}
    \label{grasp}
    \vskip -0.1in
\end{table}

\section{Conclusions and Future Work} 
This paper presents a systematic vision-based method for finding kinematic mappings between the anthropomorphic robot hand and the human hand.
This method develops an end-to-end teacher-student network (TeachNet) and creates a dataset containing 400K pairs of the human hand depth images, simulated robot depth images in the same poses and a corresponding robot joint angle. 
This dataset generation method, which maps the keypoints position and link direction of the robot from the human hand by an improved mapping method, and manipulates the robot and records the robot state in simulation, is efficient and reliable. 
By the network evaluation and the robotic experiments, we verify the applicability of the Teach Hard-Late method to model poses and the implicit correspondences between robot imitators and human demonstrators. 
The experimental results also present that our end-to-end teleoperation allows novice teleoperators to grasp the in-hand objects faster and more accurately than the HandIK solution.

Although our method performs well in real-world tasks, it has some limitations. First, it requires the operator's hand in a fixed range and has a higher error of occluded joints. Since 3D volumetric representation outperforms 2D input on capturing the spatial structure of the depth data, training an end-to-end model combining a higher level representation would likely lead to more efficient training.
Second, when we manipulated the robot to grasp tasks, we did not consider the precious tactile feedback of the robot. To perform more complicated robotic tasks, we are going to use the tactile modality of the Shadow hand combined with our teleoperation method. 
In addition, we would like to extend our method for teleoperating other body parts of the robots. 


\small{
\section*{ACKNOWLEDGMENT}
This research was funded jointly by the National Science Foundation of China (NSFC) and the German Research Foundation (DFG) in project Cross Modal Learning, NSFC 61621136008/DFG TRR-169. It was also partially supported by National Science Foundation of China (Grant No.91848206, U1613212) and project STEP2DYNA (691154). 
We would like to thank Chao Yang, Dr. Norman Hendrich, and Prof. Huaping Liu for their generous help and insightful advice. 
}
\bibliographystyle{IEEEtran} 
\bibliography{IEEEabrv,ref}
\end{document}